\documentclass[a4paper,fleqn]{cas-dc}

\usepackage{natbib}
\usepackage{flushend}

\newif\ifanon
\anonfalse

\ExplSyntaxOn
\RenewDocumentCommand \printorcid { }
{
  \seq_if_empty:NF \g_stm_orcid_seq
  {
    \group_begin:
      \tex_let:D \thefootnote \relax \footnotetext
      {
        \raggedright
        \textsc{orcid}(s):\c_space_token
        \seq_use:Nn \g_stm_orcid_seq { ;~ }
      }
    \group_end:
  }
}
\ExplSyntaxOff

\ExplSyntaxOn
\cs_new:Npn \camerl_fn_symbol:n #1
  { \int_case:nn { #1 } { { 1 } { $\ast$ } { 2 } { $\dagger$ } { 3 } { $\ddagger$ } } }

\cs_set:cpn { process@marks }
{
  \cs_if_free:cTF { mark@corau\theauthor }
    { \ignorespaces }
    { \str_set:Nx \l_tmpa_str { \use:c{ mark@corau\theauthor } }
      \int_case:nn { \l_tmpa_str }
        {
          { 1 } { \sep$\ast$ }
          { 2 } { \sep$\ast\ast$ }
          { 3 } { \sep$\ast\!\ast\!\ast$ }
        }
        \tex_def:D \sep{\unskip,}
      }
   \cs_if_free:cTF { mark@fnau\theauthor }
     { \ignorespaces }
     { \sep \camerl_fn_symbol:n { \use:c { mark@fnau\theauthor } }
       \tex_def:D \sep{\unskip,}
     }
}
\ExplSyntaxOff

\makeatletter
\def\@dbflt#1{\@ifnextchar[{\@xdblfloat{#1}}%
  {\edef\reserved@a{\noexpand\@xdblfloat{#1}[\csname fps@#1\endcsname]}\reserved@a}}
\makeatother

\begin{document}
\let\WriteBookmarks\relax
\renewcommand\floatpagefraction{.5}
\renewcommand\dblfloatpagefraction{.5}
\renewcommand\textfraction{.001}

% Short title and short author for running heads
\shorttitle{CMRL: Collision-Aware and Memory-Enhanced RL for UAV Navigation}
\ifanon
\shortauthors{Anonymous}
\else
\shortauthors{H. Hong et al.}
\fi

\title[mode=title]{CMRL: Collision-Aware and Memory-Enhanced Reinforcement Learning for UAV Navigation in Multi-Scale Obstacle Environments}

\ifanon
\else
\author[1,2]{Hong Hong}
\fnmark[1]

\author[1,2]{Feiyu Liao}
\fnmark[1]

\author[1,2]{Yongheng Liang}

\author[1,2]{Boning Zhang}

\author[3]{Michael Kunyu Wang}

\author[1,2]{Haitao Wang}
\fnmark[2]

\author[1,2]{Hejun Wu}

\affiliation[1]{organization={School of Computer Science and Engineering, Sun Yat-sen University},
            city={Guangzhou},
            country={China}}

\affiliation[2]{organization={Guangdong Key Laboratory of Big Data Analysis and Processing},
            city={Guangzhou},
            country={China}}

\affiliation[3]{organization={The Hong Kong Polytechnic University},
            city={Hong Kong},
            country={China}}

\nonumnote{\hspace*{-1.8em}$^{*}$ Equal contribution.\quad $^{\dagger}$ Corresponding author.\\
\hspace*{1.1em}E-mail address: wanght76@mail.sysu.edu.cn (H. Wang).}
\fi

%%%%%%%%%%%%%%%%%%%%%%%%%%%%%%%%%%%%%%%%%%%%%%%%%%%%%%%%%%%%%%%%%%%%%%%%%%%%%%%%
\begin{abstract}
In obstacle avoidance navigation of unmanned aerial vehicles (UAVs),
variations in obstacle scale have received less attention than obstacle number or density.
Existing methods typically extract purely geometric features from single-frame depth observations.
Such representations tend to neglect small obstacles and lose spatial context under occlusions caused by large obstacles,
leading to noticeable degradation in environments with multi-scale obstacles.
To address this issue, we propose CMRL,
a Collision-aware and Memory-enhanced Reinforcement Learning framework for UAV navigation.
The collision-aware latent representation encodes risk-sensitive depth cues to preserve fine-grained obstacle structures,
thereby improving sensitivity to small obstacles.
The temporal memory module integrates observations across frames, mitigating partial observability caused by large-obstacle occlusions.
We evaluate CMRL with multi-scale obstacles, including ultra-small and extra-large obstacle settings.
Results show that CMRL outperforms state-of-the-art baselines across all scales,
with success rate gains of 0.47 and 0.29 in the ultra-small and extra-large settings, respectively.
More importantly, CMRL achieves reliable navigation in cluttered outdoor environments.
\ifanon
The code is available at \url{https://anonymous.4open.science/r/artifact-B8E5/}.
\else
The code is available at \url{https://honghongdev.github.io/camerl/}.
\fi
\end{abstract}

% Keywords

\maketitle

%%%%%%%%%%%%%%%%%%%%%%%%%%%%%%%%%%%%%%%%%%%%%%%%%%%%%%%%%%%%%%%%%%%%%%%%%%%%%%%%
\section{Introduction}

Unmanned aerial vehicles (UAVs) are increasingly deployed in infrastructure inspection, forest surveying, search and rescue, and logistics delivery \citep{outay2020applications,wang2023cooperative}.
In these missions the vehicle has to fly close to the structures it observes, so autonomous navigation in cluttered space becomes a basic requirement \citep{xiao2025vision}.
The obstacles encountered in a single mission rarely share a common size.
A power line inspection flight faces conductors of a few centimeters together with pylons several meters wide \citep{xu2022power}.
A forest survey flight faces slender branches together with thick trunks \citep{liu2022large}.
The obstacle size within one mission can therefore span two orders of magnitude.
A practical deployment needs one onboard policy that covers this whole range under a limited computing budget \citep{kulkarni2024reinforcement}, instead of a separate model tuned for each type of scene.

Conventional UAV navigation follows a sequential pipeline of perception, mapping, planning, and control \citep{wang2023learning,zhou2020ego,zheng2024fast}.
The strong coupling between modules can cause error accumulation and imposes a substantial onboard computational burden \citep{loquercio2021learning,zhou2020ego}.
The mapping stage also represents the environment at a fixed spatial resolution \citep{zhou2020ego}, which already embeds an assumption about the obstacle size the system expects.
Learning-based methods offer a more reactive alternative by mapping onboard observations directly to control commands \citep{loquercio2021learning,nguyen2022motion}.
Reinforcement learning (RL) trained in high-fidelity simulation has shown strong performance in complex environments \citep{song2023reaching,kaufmann2023champion}.

Most studies on UAV obstacle avoidance characterize scene difficulty by the number of obstacles and by their spatial density \citep{bhattacharya2025vision,zhao2024learning}.
Obstacle scale receives much less attention.
Training environments and benchmarks are usually populated with obstacles of a similar size \citep{yu2023avoidbench}.
A policy is therefore rarely required to adapt its behavior to a change of scale.
This is a convenient simplification, but it does not match the missions described above.
As we show in Section~\ref{sec:experiments}, a policy tuned to nominal-scale scenes degrades sharply once obstacles of very different sizes appear together.

Obstacle scale affects navigation through two distinct failure modes.
Small obstacles occupy few pixels in a depth image and are further weakened by sensor noise and by the compression applied in learned representations \citep{kulkarni2023semantically}.
A policy that misses them collides without ever perceiving the cause.
Avoiding this failure requires perceptual cues that stay fine-grained and remain tied to collision risk \citep{ren2025safety,xu2025flying}.
Large obstacles create the opposite problem.
A single large obstacle can fill most of the field of view and hide the free space behind it \citep{lee2025quadrotor,zhai2025pa}.
A policy that reacts only to the current frame then oscillates or reverses its detour direction \citep{ni2022recurrent}.
Avoiding this failure requires spatial context accumulated beyond the current observation \citep{yu2025mavrl}.
The two requirements pull in opposite directions.
Preserving detail favors a faithful single-frame representation, whereas maintaining context favors aggregation across frames.
Existing methods therefore address one end of this spectrum and rarely both \citep{kulkarni2023semantically,chen2025general}.

To meet both requirements within one framework, we propose CMRL, a Collision-aware and Memory-enhanced Reinforcement Learning framework for UAV visual navigation.
A collision-aware branch converts each raw depth image into a representation that encodes the collision risk implied by the UAV body size, so that thin structures survive latent compression.
A temporal memory branch aggregates the latent codes over a fixed temporal interval, so that free space observed before an occlusion remains available to the policy.
Both modules are trained before the policy and are frozen during policy learning, which keeps reinforcement learning in a low-dimensional latent space.

We evaluate CMRL on AvoidBench \citep{yu2023avoidbench} in six test environments whose obstacle diameter ranges from 1 cm to 500 cm.
Training uses only a single nominal-scale environment, and the ultra-small, large and extra-large ranges lie entirely outside the training diameter range.
CMRL reaches the highest success rate in every environment.
Compared with the strongest baseline, the success rate improves by 0.47 in the ultra-small setting and by 0.29 in the extra-large setting.
The advantage holds across obstacle densities.
Outdoor flights in a forest containing both tree trunks and slender poles confirm that the policy transfers to a physical platform.

Our main contributions are as follows:
\begin{itemize}
  \item We formulate multi-scale obstacle avoidance as a research problem for learning-based UAV navigation.
  We build six test environments in which the obstacle diameter spans two orders of magnitude, from ultra-small to extra-large.
  We show that existing methods lose a large part of their success rate at both extremes of this range.
  \item We propose CMRL, which addresses the two scale-dependent requirements with two dedicated modules.
  Collision-aware representation learning preserves fine structures and encodes collision risk in the latent space.
  Temporal memory aggregates observations to recover the spatial context lost to occlusion.
  \item We evaluate CMRL through ablation studies, baseline comparisons, density sweeps, and attention visualization, and we deploy the learned policy on a physical quadrotor.
  The results show consistent gains at every obstacle scale and reliable flight in a cluttered outdoor forest.
  The training and evaluation code is publicly available.
\end{itemize}

The remainder of this paper is organized as follows.
Section~\ref{sec:related} reviews related work on learning-based visual navigation and on obstacle avoidance in multi-scale environments.
Section~\ref{sec:representation} presents the collision-aware and memory-enhanced representation learning framework.
Section~\ref{sec:rl} describes the reinforcement learning formulation for navigation.
Section~\ref{sec:experiments} reports the simulation and real-world results.
Section~\ref{sec:conclusion} concludes the paper.

\section{Related Work}
\label{sec:related}

\begin{figure*}[pos=t]
  \centering
  \includegraphics[width=\textwidth]{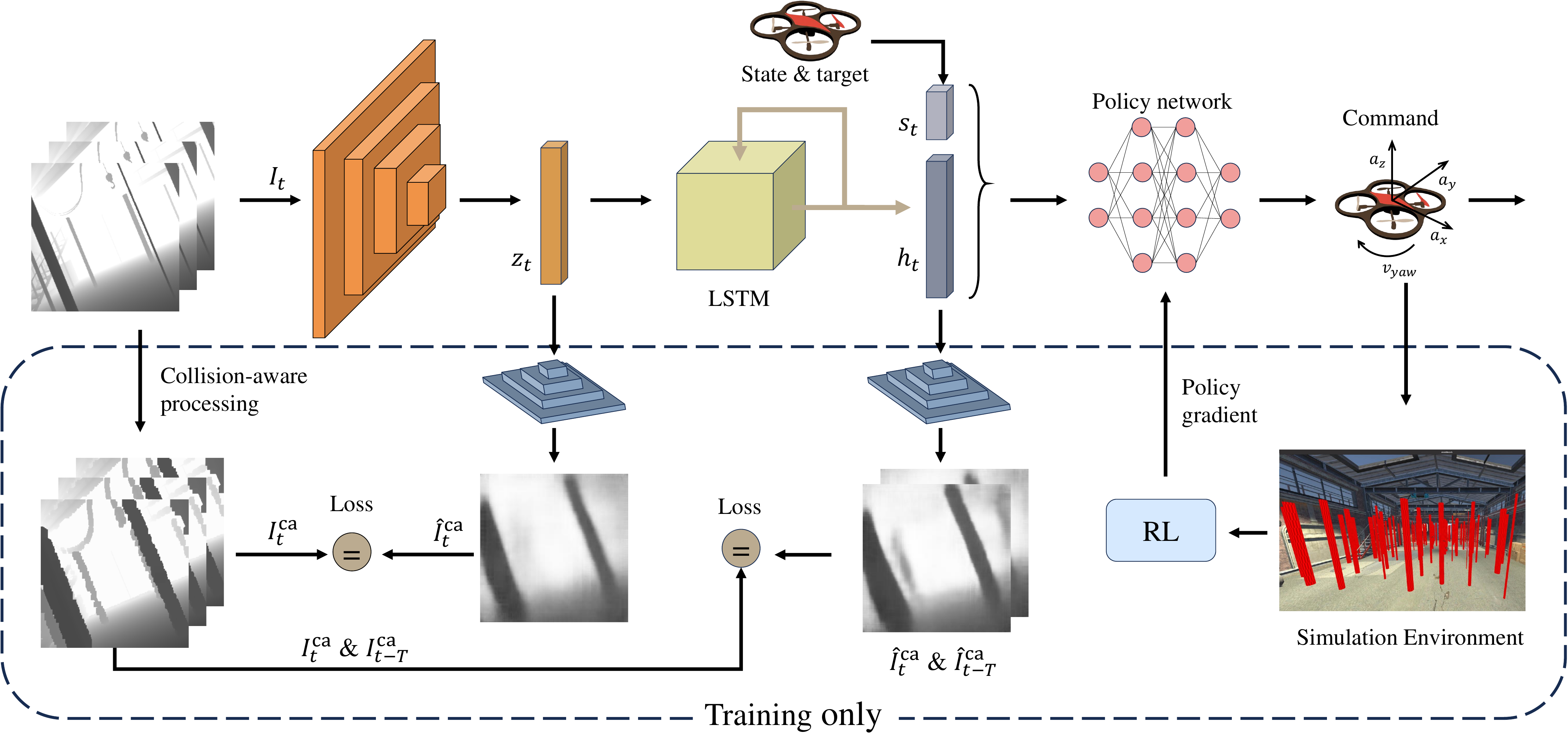}
  \caption{Overview of the CMRL architecture and training pipeline.}
  \label{fig:architecture}
\end{figure*}

\subsection{Learning-Based Visual Navigation}
Learning-based visual navigation is commonly studied through imitation learning and reinforcement learning.
Imitation learning directly maps visual observations to control commands using expert demonstrations, but its performance is often limited by the expert policy and may generalize poorly in complex environments \citep{loquercio2021learning,bhattacharya2025vision}.
\citet{kim2025rapid} instead recover a cost function from expert trajectories through inverse reinforcement learning, which yields a planner that is less tied to the demonstrated behavior.
Reinforcement learning instead optimizes long-term returns through interaction and can better balance safety and efficiency, although end-to-end visual RL often suffers from sample inefficiency and instability \citep{kaufmann2023champion}.

To improve training efficiency and robustness, many recent methods adopt a two-stage pipeline that first learns compact visual representations and then trains policies in the latent space.
\citet{hoeller2021learning} fuse sequential depth images with camera trajectories to learn local structural representations for dynamic obstacle avoidance.
\citet{kulkarni2024reinforcement} compress depth observations into a compact latent encoding that is combined with odometry and goal information to train a low-latency navigation policy with robust sim-to-real transfer.
\citet{yu2025mavrl} learn an adaptive flight-speed policy that adjusts velocity to the surrounding environment complexity, and further bridge the gap between simulated and real depth inputs through a domain-adaptation-based transfer method for robust sim-to-real deployment \citep{yu2025depth}.
Their latent representation reconstructs both the current and a past depth observation to retain a short-term memory of the surrounding structure, using the raw depth image as the reconstruction target.
For RGB navigation, \citet{zhang2025learning} learn depth-consistent RGB representations through cross-modal contrastive learning.

Onboard observations cover only part of the scene, so visual navigation is naturally a partially observable decision problem.
A common remedy is to give the policy a recurrent state that summarizes past observations.
\citet{ni2022recurrent} show that recurrent model-free reinforcement learning is a strong baseline across a wide range of partially observable tasks.
\citet{xue2023multi} apply a recurrent network to multi-agent UAV navigation in unknown complex environments, where each vehicle acts on incomplete knowledge of its surroundings.
These methods aggregate past information to compensate for a limited field of view in general.
CMRL instead aggregates collision-aware latent codes, and it targets the specific partial observability caused by large obstacles.

Despite these advances, most existing methods assume obstacles of similar scales and rarely study how scale variations affect perception and policy learning. In contrast, CMRL explicitly targets multi-scale settings by coupling collision-aware representation learning with temporal memory.

\subsection{Obstacle Avoidance in Multi-Scale Environments}
In multi-scale environments, existing methods often address small and large obstacles separately.
Small obstacles are easily missed in visual or depth observations, motivating methods that enhance geometric representations or safety modeling for more reliable avoidance.
For example, \citet{kulkarni2023semantically} learn a semantically enhanced depth representation that preserves slender structures under compression,
while \citet{ren2025safety} and \citet{xu2025flying} improve avoidance of tiny obstacles and narrow passages using fine-grained LiDAR-based planning or reinforcement learning.

Large obstacles mainly introduce severe occlusions and detouring difficulty, which motivates approaches based on path generation or privileged information.
\citet{chen2025general} improve path reliability near large obstacles through local sub-goals and a backtracking detour mechanism.
\citet{lee2025quadrotor} use privileged cues to maintain the correct detouring direction under occlusion.
\citet{zhai2025pa} propose perception-aware model predictive path integral (MPPI) control to encourage exploration when the goal is occluded.
However, jointly handling small-obstacle safety and large-obstacle occlusion in coexisting multi-scale environments remains underexplored. CMRL addresses this gap with a unified framework that combines collision-aware representation learning and temporal memory.

\section{Collision-Aware and Memory-Enhanced Representation Learning}
\label{sec:representation}

\begin{figure*}[pos=t]
  \centering
  \includegraphics[width=\textwidth]{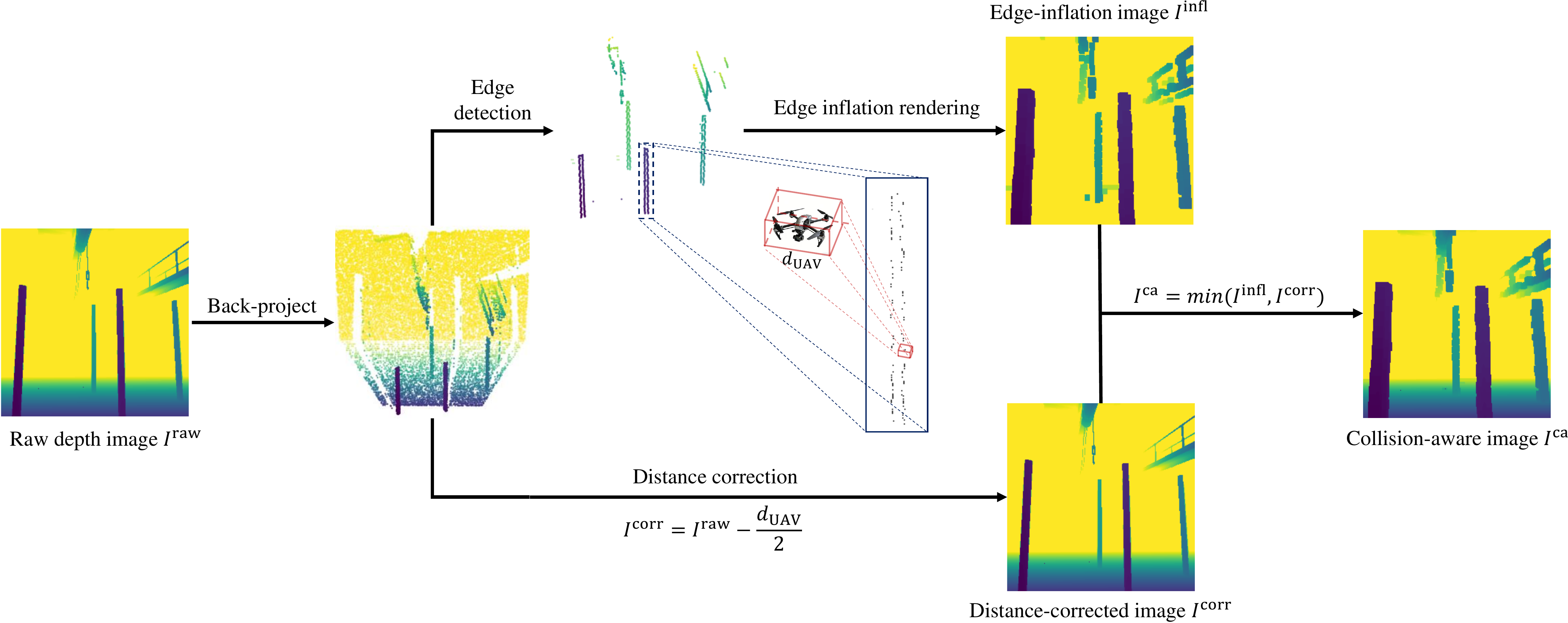}
  \caption{Collision-aware preprocessing pipeline.}
  \label{fig:3_ca_processing}
\end{figure*}

This section presents CMRL, which realizes collision-aware and memory-enhanced representation learning for multi-scale obstacle avoidance.
As illustrated in Fig.~\ref{fig:architecture}, CMRL encodes each depth image with a variational autoencoder (VAE) and fuses temporal observations through a long short-term memory (LSTM) network to obtain a memory-enhanced representation.
This representation is concatenated with the UAV state and target information for policy prediction.
The training pipeline proceeds in three stages.
We first train an initial policy to collect depth-image sequences, then train the VAE and LSTM to learn collision-aware and memory-enhanced representations, and finally freeze the representation modules and retrain the policy network.

\subsection{Dataset Collection and Collision-Aware Processing}

To build the training dataset, we first train an initial policy while keeping the VAE and LSTM randomly initialized and fixed.
This policy mainly learns basic goal-reaching behavior and is used to collect depth-image sequences in the nominal-scale training environment.
The representation modules therefore never observe the extreme obstacle scales used for evaluation.
Because the dataset is gathered from continuous trajectories, the resulting observations naturally contain temporal correlations, which are useful for subsequent LSTM training.

After obtaining the raw depth-image dataset, we perform collision-aware preprocessing to account for the UAV body size and generate collision-aware depth images, as illustrated in Fig.~\ref{fig:3_ca_processing}.
Given a raw depth image $I^{\mathrm{raw}}$, we first extract obstacle contours and back-project them into 3D.
The contour regions are then inflated according to the effective UAV size $d_{\mathrm{UAV}}$, and rendered to produce an edge-inflation image $I^{\mathrm{infl}}$, which emphasizes collision risks near obstacle boundaries.
For obstacle interiors, we apply distance correction by subtracting half of the UAV size from the raw depth image, yielding $I^{\mathrm{corr}} = I^{\mathrm{raw}} - d_{\mathrm{UAV}}/2$.
The final collision-aware depth image is then obtained by pixel-wise fusion, $I^{\mathrm{ca}} = \min\left(I^{\mathrm{infl}}, I^{\mathrm{corr}}\right)$.
This result provides collision-aware supervision for the subsequent representation learning stage.

\subsection{Collision-Aware Representation Learning}
After obtaining paired raw depth images $I_t^{\mathrm{raw}}$ and collision-aware depth images $I_t^{\mathrm{ca}}$, we train a VAE to learn collision-aware latent representations from $I_t^{\mathrm{raw}}$.
The VAE encoder comprises six convolutional layers followed by two fully connected layers, which output the latent mean and variance.
Symmetrically, the decoder employs six deconvolutional layers and concludes with a Sigmoid activation.
The latent dimension is set to $n_z = 64$.
During training, the encoder maps $I_t^{\mathrm{raw}}$ to the mean $\mu_t$ and standard deviation $\sigma_t$ of a Gaussian latent distribution, from which a latent vector $z_t$ is sampled via the reparameterization trick:
\begin{equation}
z_t \sim \mathcal{N}\!\left( \mu_t,\; \mathrm{diag}(\sigma_t^2) \right).
\end{equation}

The decoder reconstructs the corresponding collision-aware image $\hat{I}_t^{\mathrm{ca}}$ from $z_t$, using $I_t^{\mathrm{ca}}$ as the supervision target.
This forces the latent space to preserve geometric structures and safety-related cues essential for collision reasoning.
This design embeds collision-aware information without requiring preprocessing at inference time, ensuring lightweight onboard computation.

The training objective consists of a reconstruction loss $\mathcal{L}_{\mathrm{coll}}$ computed via the mean squared error (MSE), and a Kullback-Leibler (KL) divergence regularization term $\mathcal{L}_{\mathrm{KL}}$.
The overall loss $\mathcal{L}_{\mathrm{CA}}$ is thus defined as:
\begin{equation}
\begin{aligned}
\mathcal{L}_{\mathrm{CA}} &= \mathcal{L}_{\mathrm{coll}} + \lambda_{\mathrm{KL}}\,\mathcal{L}_{\mathrm{KL}},\\
\mathcal{L}_{\mathrm{coll}} &= \mathrm{MSE}\!\left(I_t^{\mathrm{ca}},\,\hat{I}_t^{\mathrm{ca}}\right),\\
\mathcal{L}_{\mathrm{KL}} &= -\frac{1}{2}\sum_{i=1}^{n_z}
\left(1+\log(\sigma_{t,i}^2)-\mu_{t,i}^2-\sigma_{t,i}^2\right),
\end{aligned}
\end{equation}

where $\lambda_{\mathrm{KL}}$ is the weight of the KL regularization term, $n_z$ denotes the latent dimensionality, $\mu_{t,i}$ and $\sigma_{t,i}$ are the $i$-th components of $\mu_t$ and $\sigma_t$,
and $I_t^{\mathrm{ca}}$ and $\hat{I}_t^{\mathrm{ca}}$ denote the ground-truth and reconstructed collision-aware images, respectively.

\subsection{Memory-Enhanced Representation Learning}
Using the pretrained encoder, each raw depth observation $I_t^{\mathrm{raw}}$ is first mapped to a latent vector $z_t$.
The sequence of latent vectors is then processed by a single-layer LSTM with a 256-dimensional hidden state $h_t$ that summarizes temporal context.
A prediction head maps $h_t$ to two predicted latent codes $\tilde z_{t-T}$ and $\tilde z_t$, corresponding to a past frame at time $t-T$ and the current frame at time $t$, where $T$ denotes the temporal interval.
These codes are then decoded by the frozen VAE decoder into the reconstructed collision-aware images $\hat{I}_{t-T}^{\mathrm{ca}}$ and $\hat{I}_{t}^{\mathrm{ca}}$.
By jointly reconstructing $\hat{I}_{t-T}^{\mathrm{ca}}$ and $\hat{I}_{t}^{\mathrm{ca}}$, the supervision encourages $h_t$ to retain temporally consistent scene structure and collision-relevant information,
which is especially useful when the current observation is partially occluded by large obstacles.

The LSTM is trained by minimizing the sum of reconstruction errors over the two time steps:
\begin{equation}
\mathcal{L}_{\mathrm{LSTM}}=
\mathrm{MSE}\!\left(I_{t}^{\mathrm{ca}},\,\hat{I}_{t}^{\mathrm{ca}}\right)
+
\mathrm{MSE}\!\left(I_{t-T}^{\mathrm{ca}},\,\hat{I}_{t-T}^{\mathrm{ca}}\right).
\end{equation}
Minimizing this objective yields a memory-enhanced representation $h_t$ for downstream policy learning.
This temporal reconstruction scheme follows \citet{yu2025mavrl}, with the collision-aware image replacing the raw depth image as the reconstruction target.
Table~\ref{tab:repr_eval} quantifies this change through the comparison between MeRL and CMRL.

\section{Reinforcement Learning for Navigation}
\label{sec:rl}
\subsection{Task Formulation}

We formulate UAV navigation as a partially observable Markov decision process (POMDP). The policy
$\pi_\theta(a_t\mid o_t)$, implemented as a multilayer perceptron, is trained with Proximal Policy Optimization (PPO) to maximize the expected discounted return under partial observations.
The action $a_t$ is defined as the high-level control command $[a_x,a_y,a_z,v_{\mathrm{yaw}}]$, where
$[a_x,a_y,a_z]$ denotes the desired acceleration and $v_{\mathrm{yaw}}$ denotes the yaw rate.
The observation $o_t$ concatenates the UAV state and target information $s_t$ with the
memory-enhanced representation $h_t$, giving $o_t=[s_t,h_t]$. Specifically,
\begin{equation}
s_t=\big[\log d_{\mathrm{hor}},\ d_z,\ d_{\mathrm{norm}},\ v^{\mathrm{W}},\ e,\ \omega^{\mathrm{B}}\big],
\end{equation}
where $d_{\mathrm{hor}}$ and $d_z$ denote the horizontal and vertical distances to the target,
$d_{\mathrm{norm}}$ is the normalized target direction, $v^{\mathrm{W}}$ is the linear velocity in the
world frame, $e$ denotes the Euler-angle attitude, and $\omega^{\mathrm{B}}$ is the body-frame angular velocity.

\subsection{Reward Function}
We design the reward $r_t$ as the combination of terminal rewards and a dense progress reward. An episode terminates when the UAV reaches the goal, collides with obstacles, or leaves the allowed flight region. The overall reward at time step $t$ is defined as
\begin{equation}
r_t=
\begin{cases}
r_{\mathrm{arrive}}, & \|d_t\|<d_{\min},\\
r_{\mathrm{collision}}, & \text{a collision occurs},\\
r_{\mathrm{exceed}}, & p_t \notin [p_{\min},\,p_{\max}],\\
r_{\mathrm{prog}}, & \text{otherwise},
\end{cases}
\end{equation}
where $r_{\mathrm{arrive}}$, $r_{\mathrm{collision}}$, and $r_{\mathrm{exceed}}$ are constant terminal rewards for goal arrival, collision, and leaving the flight region, respectively, while $r_{\mathrm{prog}}$ is the dense progress reward defined below. $d_t$ denotes the 3D relative displacement vector to the goal, so that $\|d_t\|=\sqrt{d_{\mathrm{hor}}^2+d_z^2}$, and $p_t=[x_t,y_t,z_t]$ is the UAV position. $d_{\min}$ is the arrival threshold, and $[p_{\min},p_{\max}]$ denotes the valid flight region defined by the lower and upper bounds along each axis.

The progress reward is defined as
\begin{equation}
\begin{aligned}
r_{\mathrm{prog}}=&\ \lambda_d \log d_{\mathrm{hor}} + \lambda_z d_z
+ \lambda_v\,\max(0, v_{\mathrm{hor}}-v_{\max})\,v_{\mathrm{hor}} \\
&+ \lambda_{\mathrm{dir}}\left\|d_{\mathrm{norm}}-v_{\mathrm{norm}}\right\|_1
+ \lambda_{\mathrm{ang}}\left\|\omega^{\mathrm{B}}\right\| \\
&+ \lambda_{\mathrm{lat}}\left(|v_y^{\mathrm{B}}|+\max(0,-v_x^{\mathrm{B}})\right),
\end{aligned}
\end{equation}
where $\lambda_d$, $\lambda_z$, $\lambda_v$, $\lambda_{\mathrm{dir}}$, $\lambda_{\mathrm{ang}}$, and $\lambda_{\mathrm{lat}}$ are weighting coefficients for the individual reward terms.
$d_{\mathrm{hor}}$ and $d_z$ are the horizontal and vertical distances to the goal,
$v_{\mathrm{hor}}$ is the horizontal speed with limit $v_{\max}$, $d_{\mathrm{norm}}$ and
$v_{\mathrm{norm}}$ are the normalized goal and velocity directions, $\omega^{\mathrm{B}}$ is the body-frame
angular velocity, and $v_x^{\mathrm{B}}$, $v_y^{\mathrm{B}}$ are the forward and lateral body-frame velocities.
The distance and direction terms encourage goal reaching, while the remaining terms regularize
speed, angular motion, and lateral/backward movements for stable and safe flight.

\begin{figure*}[pos=b]
  \centering
  \includegraphics[width=\textwidth]{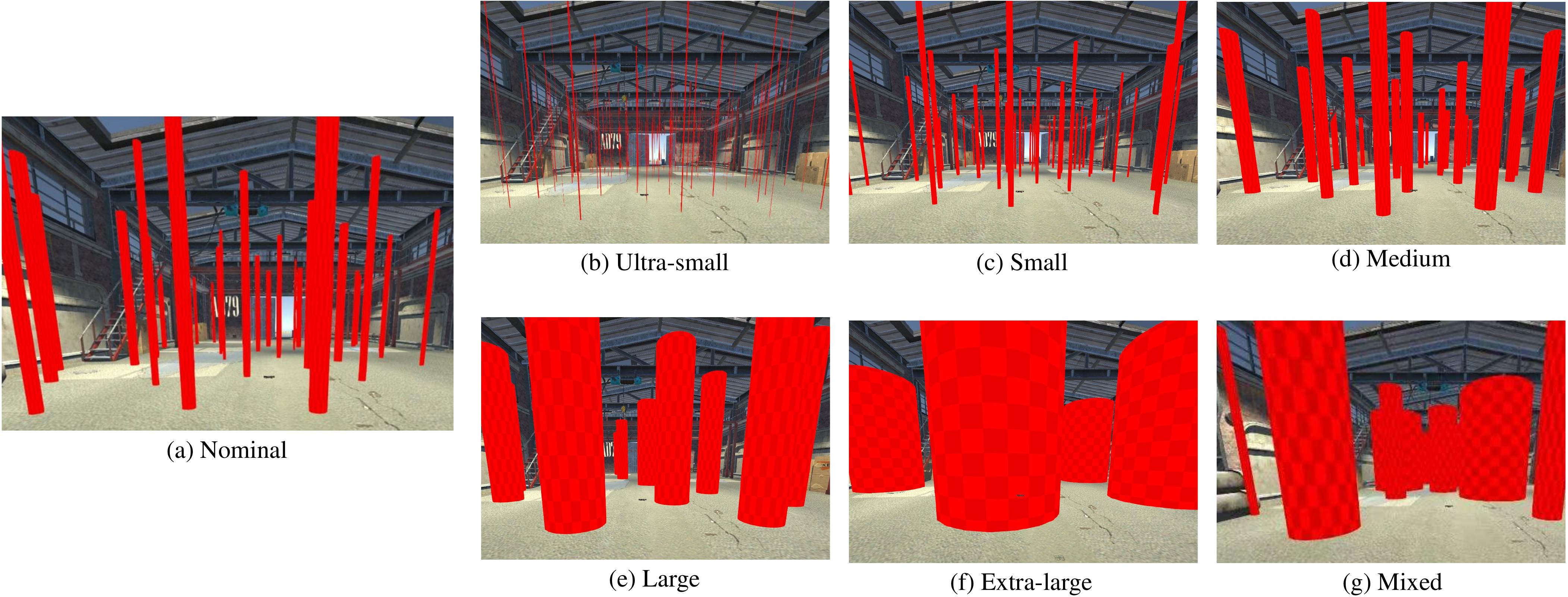}
  \caption{Simulation environments with different obstacle scales.
  (a) Nominal-scale training environment.
  (b)--(g) Test environments with ultra-small (US), small (S), medium (M), large (L), extra-large (XL), and mixed-scale (MIX) obstacles.}
  \label{fig:sim_env}
\end{figure*}
\section{Experiments}
\label{sec:experiments}

\subsection{Simulation Setup}
\label{subsec:sim_setup}

\begin{table}[t]
  \centering
  \caption{Reward function parameters.}
  \label{tab:reward_params}
  \small
  \setlength{\tabcolsep}{6pt}
  \renewcommand{\arraystretch}{1.10}
  \begin{tabular}{llr}
    \toprule
    Parameter & Description & Value \\
    \midrule
    $r_{\mathrm{arrive}}$    & Goal arrival                & $0$ \\
    $r_{\mathrm{collision}}$ & Collision                   & $-10.0$ \\
    $r_{\mathrm{exceed}}$    & Leaving the flight region   & $-10.0$ \\
    $\lambda_d$              & Horizontal distance         & $-0.01$ \\
    $\lambda_z$              & Vertical distance           & $-0.002$ \\
    $\lambda_v$              & Speed limit                 & $-0.001$ \\
    $\lambda_{\mathrm{dir}}$ & Direction alignment         & $-0.003$ \\
    $\lambda_{\mathrm{ang}}$ & Angular velocity            & $-0.005$ \\
    $\lambda_{\mathrm{lat}}$ & Lateral and backward motion & $-0.0005$ \\
    $v_{\max}$               & Horizontal speed limit      & $3.0$ m/s \\
    \bottomrule
  \end{tabular}
\end{table}

Table~\ref{tab:reward_params} lists the values used in all experiments.
Goal arrival carries no explicit bonus, so the policy is driven toward the goal by the distance terms and by the early termination of the episode.

We build the simulation environment on Flightmare \citep{song2021flightmare} with a simplified reinforcement-learning dynamics model.
To evaluate robustness to obstacle-scale variation, the representation modules and the policy are both trained in a single nominal-scale environment and then tested in six environments with different obstacle sizes, as shown in Fig.~\ref{fig:sim_env}.

The scene is an indoor garage populated with cylindrical obstacles. We construct seven environment configurations by varying obstacle diameter:
one nominal-scale training environment (10--50 cm) and six test environments, namely ultra-small (US, 1--5 cm), small (S, 10--30 cm), medium (M, 40--80 cm), large (L, 100--200 cm), extra-large (XL, 400--500 cm),
and mixed-scale (MIX, 1--500 cm). Obstacle density is controlled by the radius parameter used for Poisson obstacle placement, yielding different clutter levels and navigation difficulty.
We fix the obstacle shape to vertical cylinders so as to isolate obstacle scale as the single controlled variable.
For planar navigation, we assume that avoidance difficulty is dominated by the angular size an obstacle subtends and by the occlusion it induces.
Cylinders of varying diameter already span this range, from easily missed thin structures to strong occluders.

For representation learning, we collect trajectory data in the nominal-scale training environment using the initial exploration policy.
The resulting dataset contains 200 episodes and approximately 10,000 depth images for subsequent representation learning and model training.
No data from the six test environments is used to train the representation modules or the policy.
All cross-scale results reported below are thus obtained without adaptation to the test scale.

All simulation results reported below follow the same evaluation protocol, which is built on AvoidBench \citep{yu2023avoidbench}.
Every method is evaluated in the six test environments under densely cluttered and randomly generated layouts.
For each test environment, the Poisson disk sampling radius is varied over a range that scales with the obstacle diameter, so that traversable gaps remain available at every obstacle scale.
Each configuration is repeated eight times, and each repetition contains 500 independent flights.
Every repetition regenerates the obstacle layout, so the variation across repetitions reflects layout randomness and not only sampling noise.
A flight is counted as a success when the UAV reaches the goal without collision.
The average speed is computed over the successful flights only.
Success rates are averaged over the radius range unless a specific radius is given.
All methods share the same observation setting, termination conditions, and environment generation procedure.

\subsection{Representation Analysis}

\begin{table}[t]
  \centering
  \caption{Effect of temporal interval $T$.}
  \label{tab:t_sensitivity}
  \setlength{\tabcolsep}{4pt}
  \renewcommand{\arraystretch}{1.05}
  \begin{tabular*}{\columnwidth}{@{\extracolsep{\fill}}c cccccc@{}}
    \toprule
    $T$ & 1 & 5 & 10 & 15 & 20 & 30 \\
    \midrule
    Success rate & 0.62 & 0.76 & \textbf{0.81} & 0.73 & 0.71 & 0.68 \\
    \bottomrule
  \end{tabular*}
\end{table}

We first analyze the effect of the temporal interval $T$ in the memory module under mixed-scale obstacle settings.
Considering the limited onboard computational resources, we set the high-level control frequency to 10 Hz, and we set $T=10$ so that the memory spans a one-second history window at this frequency.
Each setting is evaluated with the protocol of Section~\ref{subsec:sim_setup}, restricted to the mixed-scale environment.
Table~\ref{tab:t_sensitivity} shows that the success rate rises from 0.62 at $T=1$ to 0.81 at $T=10$.
It then declines gradually to 0.68 at $T=30$.
The three settings between $T=5$ and $T=15$ stay within 0.08 of each other, so the policy is not sensitive to the exact value in this range.
This trend indicates that a moderate temporal interval provides useful historical context, whereas excessively short or long intervals are less effective.
The analysis confirms that this choice sits in the stable region, and we use $T=10$ in all following experiments.

\begin{table*}[pos=tb]
  \centering
  \caption{Representation analysis across obstacle scales.}
  \label{tab:repr_eval}
  \setlength{\tabcolsep}{5pt}
  \renewcommand{\arraystretch}{1.15}
  \begin{tabular*}{\textwidth}{@{\extracolsep{\fill}}c cccc cccc@{}}
    \toprule
    \multirow{2}{*}{Environment} &
    \multicolumn{4}{c}{Success rate} &
    \multicolumn{4}{c}{Average speed (m/s)} \\
    \cmidrule(lr){2-5}\cmidrule(lr){6-9}
    & VanillaRL & CaRL & MeRL & CMRL
    & VanillaRL & CaRL & MeRL & CMRL \\
    \midrule
    Ultra-small & 0.26 & 0.63 & 0.43 & \textbf{0.77} & 1.54 & 2.05 & 1.45 & \textbf{2.08} \\
    Small       & 0.72 & 0.81 & 0.78 & \textbf{0.83} & 1.87 & 1.96 & 1.59 & \textbf{2.04} \\
    Medium      & 0.55 & 0.88 & 0.84 & \textbf{0.91} & 1.53 & \textbf{2.00} & 1.49 & 1.87 \\
    Large       & 0.48 & 0.83 & 0.81 & \textbf{0.90} & 1.48 & \textbf{2.16} & 1.42 & 2.05 \\
    Extra-large & 0.39 & 0.57 & 0.62 & \textbf{0.79} & 1.32 & \textbf{1.83} & 1.44 & 1.68 \\
    Mixed       & 0.12 & 0.73 & 0.52 & \textbf{0.81} & 0.86 & 1.69 & 1.38 & \textbf{1.89} \\
    \bottomrule
  \end{tabular*}
\end{table*}

To evaluate the contribution of collision awareness and memory enhancement, we compare four policy variants under the same reinforcement-learning framework: VanillaRL, CaRL, MeRL, and CMRL.
VanillaRL uses neither module, CaRL uses collision awareness only, MeRL uses memory enhancement only, and CMRL combines both.
All variants are evaluated with the protocol of Section~\ref{subsec:sim_setup}.

As shown in Table~\ref{tab:repr_eval}, VanillaRL performs poorly in densely cluttered multi-scale environments.
Its success rate falls to 0.12 in the mixed setting and to 0.26 in the ultra-small setting.
Adding either module raises the success rate in every environment.

The two modules are effective in different regimes.
CaRL exceeds MeRL by 0.20 in the ultra-small setting and by 0.21 in the mixed setting, which indicates improved sensitivity to fine structures and safety margins.
The extra-large setting is the only environment where MeRL exceeds CaRL.
This suggests that temporal context matters most when large obstacles occlude the field of view.
In the small, medium, and large settings the two variants stay within 0.04 of each other.

CMRL achieves the highest success rate in all six environments.
Its gain over the better of the two single-module variants is largest at the two extreme scales, with 0.14 in the ultra-small setting and 0.17 in the extra-large setting.
Neither module alone is sufficient in these two settings, whereas the combination recovers a large part of the gap.
These results suggest that collision awareness and memory enhancement play complementary roles in multi-scale obstacle avoidance.

The average speed reveals a further trade-off.
MeRL is the slowest variant in most environments, and its speed stays within the narrow band from 1.38 m/s to 1.59 m/s.
Memory enhancement alone therefore yields a conservative policy.
CaRL reaches the highest speed in the medium, large, and extra-large settings, but always at a lower success rate than CMRL.
In the extra-large setting CaRL flies at 1.83 m/s with a success rate of 0.57, while CMRL flies at 1.68 m/s with a success rate of 0.79.
CMRL thus gives up a small amount of speed for a large gain in reliability.

\subsection{Training Efficiency}

\begin{figure}[pos=t]
  \centering
  \includegraphics[width=\linewidth]{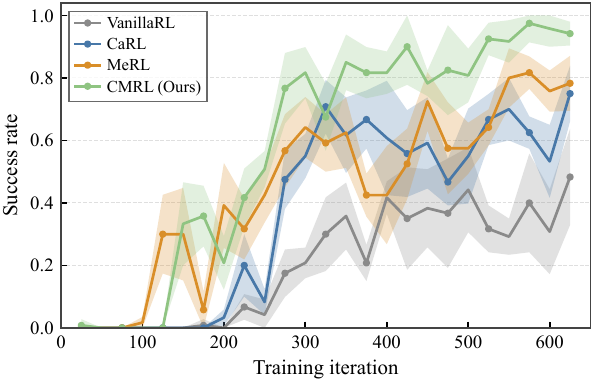}
  \caption{Training curves of the four policy variants.}
  \label{fig:train_curve}
\end{figure}

\begin{table*}[pos=b]
  \centering
  \caption{Baseline comparison across obstacle scales.}
  \label{tab:baseline_eval}
  \setlength{\tabcolsep}{5pt}
  \renewcommand{\arraystretch}{1.10}
  \begin{tabular*}{\textwidth}{@{\extracolsep{\fill}}c ccc ccc@{}}
    \toprule
    \multirow{2}{*}{Environment} &
    \multicolumn{3}{c}{Success rate} &
    \multicolumn{3}{c}{Average speed (m/s)} \\
    \cmidrule(lr){2-4}\cmidrule(lr){5-7}
    & Agile-autonomy & MAVRL & CMRL
    & Agile-autonomy & MAVRL & CMRL \\
    \midrule
    Ultra-small & 0.29 & 0.30 & \textbf{0.77} & 1.67 & 1.22 & \textbf{2.08} \\
    Small       & 0.32 & 0.46 & \textbf{0.83} & 1.72 & 1.34 & \textbf{2.04} \\
    Medium      & 0.34 & 0.71 & \textbf{0.91} & 1.70 & 1.32 & \textbf{1.87} \\
    Large       & 0.26 & 0.46 & \textbf{0.90} & 1.68 & 1.20 & \textbf{2.05} \\
    Extra-large & 0.24 & 0.50 & \textbf{0.79} & 1.64 & 1.18 & \textbf{1.68} \\
    Mixed       & 0.36 & 0.58 & \textbf{0.81} & 1.69 & 1.21 & \textbf{1.89} \\
    \bottomrule
  \end{tabular*}
\end{table*}

\begin{figure*}[pos=t]
  \centering
  \includegraphics[width=0.9\textwidth]{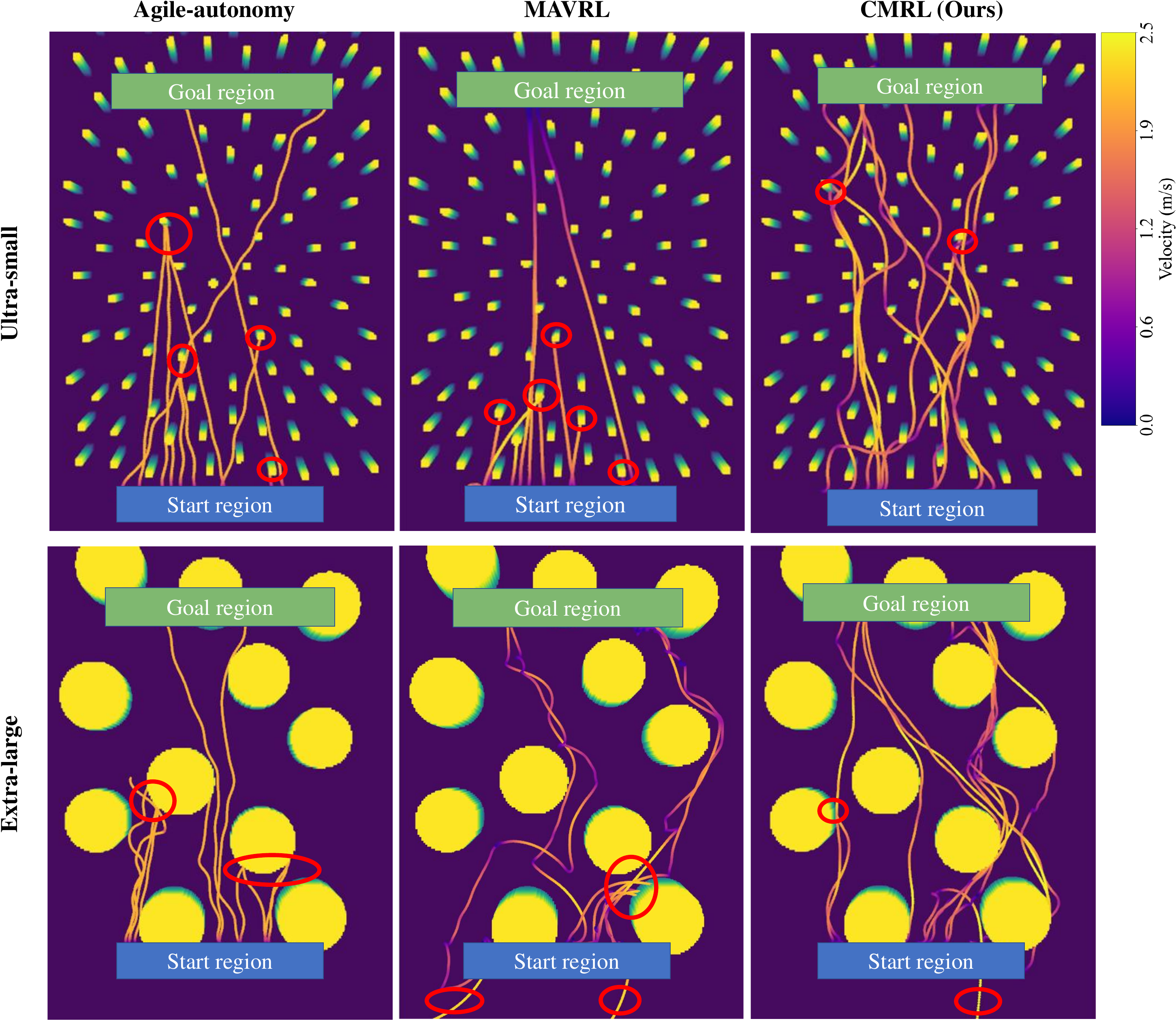}
  \caption{Representative multi-run trajectories of Agile-autonomy, MAVRL, and CMRL in extreme-scale environments.
  Red circles indicate failed trajectories, and trajectory colors denote velocity.}
  \label{fig:traj}
\end{figure*}

\begin{figure*}[pos=t]
  \centering
  \includegraphics[width=0.98\textwidth]{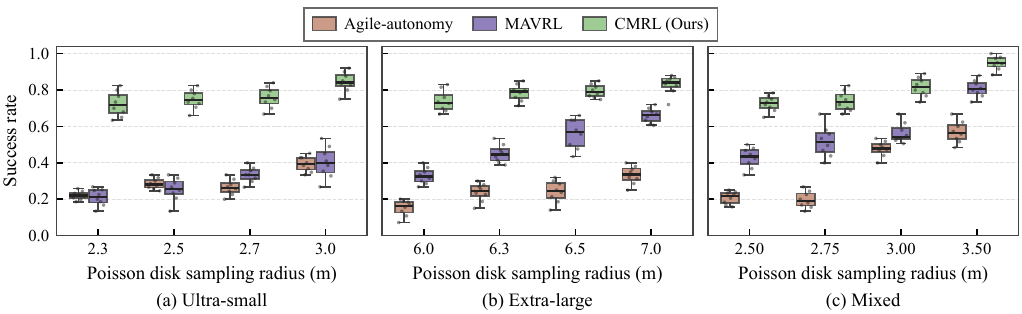}
  \caption{Success rate under different obstacle densities.
  (a) Ultra-small. (b) Extra-large. (c) Mixed.
  A smaller Poisson disk sampling radius corresponds to a denser layout.}
  \label{fig:box_density}
\end{figure*}
The comparison above is based on the final policies.
We further examine how collision awareness and memory enhancement affect the training process itself.
All four variants are trained in the nominal-scale environment with identical reinforcement-learning hyperparameters and the same number of iterations.
Every 25 iterations, the current policy is evaluated with six repetitions of 20 trials each.
We report the mean success rate over the six repetitions together with the standard deviation.

Fig.~\ref{fig:train_curve} shows the resulting training curves.
CMRL reaches a success rate of 0.77 at iteration 275 and 0.90 at iteration 425.
MeRL first exceeds the former level at iteration 550, and CaRL does not reach it within the training budget.
At the end of training, CMRL attains 0.94, while MeRL, CaRL, and VanillaRL attain 0.78, 0.75, and 0.48.
CMRL therefore reaches a given performance level with markedly fewer environment interactions.

The shaded bands reveal a similar pattern in training stability.
VanillaRL and CaRL keep wide bands until the end of training, which indicates that their success rate varies strongly across evaluation repetitions.
The band of CMRL narrows after iteration 500 and remains narrow, with a final standard deviation of 0.04 against 0.15 for VanillaRL.
These observations indicate that the two proposed modules improve sample efficiency and training stability, and not only the final navigation performance.

\subsection{Baseline Comparisons}

To evaluate CMRL against existing methods, we conduct benchmark comparisons under the protocol of Section~\ref{subsec:sim_setup}.
The compared baselines include MAVRL \citep{yu2025mavrl}, a reinforcement-learning-based method, and Agile-autonomy \citep{loquercio2021learning}, a supervised-learning-based method.
MAVRL is retrained in our simulation environment with the same number of iterations as CMRL.
Agile-autonomy is evaluated with the weights released by its authors, because its training pipeline requires privileged expert demonstrations that our setting does not provide.
Its success rate may therefore understate what a retrained model could reach.
For CMRL, the predicted acceleration commands are converted into executable body-rate and thrust commands using the model predictive control (MPC) module provided by Agilicious \citep{foehn2022agilicious}.
The CMRL results are identical to those in Table~\ref{tab:repr_eval}, since both tables report the same evaluation runs.

Table~\ref{tab:baseline_eval} summarizes the performance across obstacle scales.
CMRL achieves the highest success rate in all six environments and also maintains competitive average speed, indicating a favorable trade-off between reliability and efficiency.
The gains are particularly clear in the ultra-small and extra-large settings, which represent two extreme-scale navigation conditions.
In the ultra-small environment, CMRL improves the success rate from 0.30 to 0.77 relative to MAVRL, suggesting better sensitivity to tiny obstacles.
In the extra-large environment, CMRL improves the success rate from 0.50 to 0.79 while maintaining higher average speed, indicating stronger robustness under severe occlusions.

Fig.~\ref{fig:traj} shows representative multi-run trajectories in the ultra-small and extra-large environments.
In the ultra-small setting, Agile-autonomy and MAVRL exhibit more failed trajectories and less consistent rollouts toward the goal, whereas CMRL produces more coherent trajectories with fewer failures.
In the extra-large setting, CMRL also shows more reliable detouring behavior around large obstacles, while Agile-autonomy and MAVRL exhibit more failures and less efficient path selection.
These trajectory patterns are consistent with the quantitative results in Table~\ref{tab:baseline_eval}.

\subsection{Robustness to Obstacle Density}

\begin{figure*}[pos=t]
  \centering
  \includegraphics[width=0.9\textwidth]{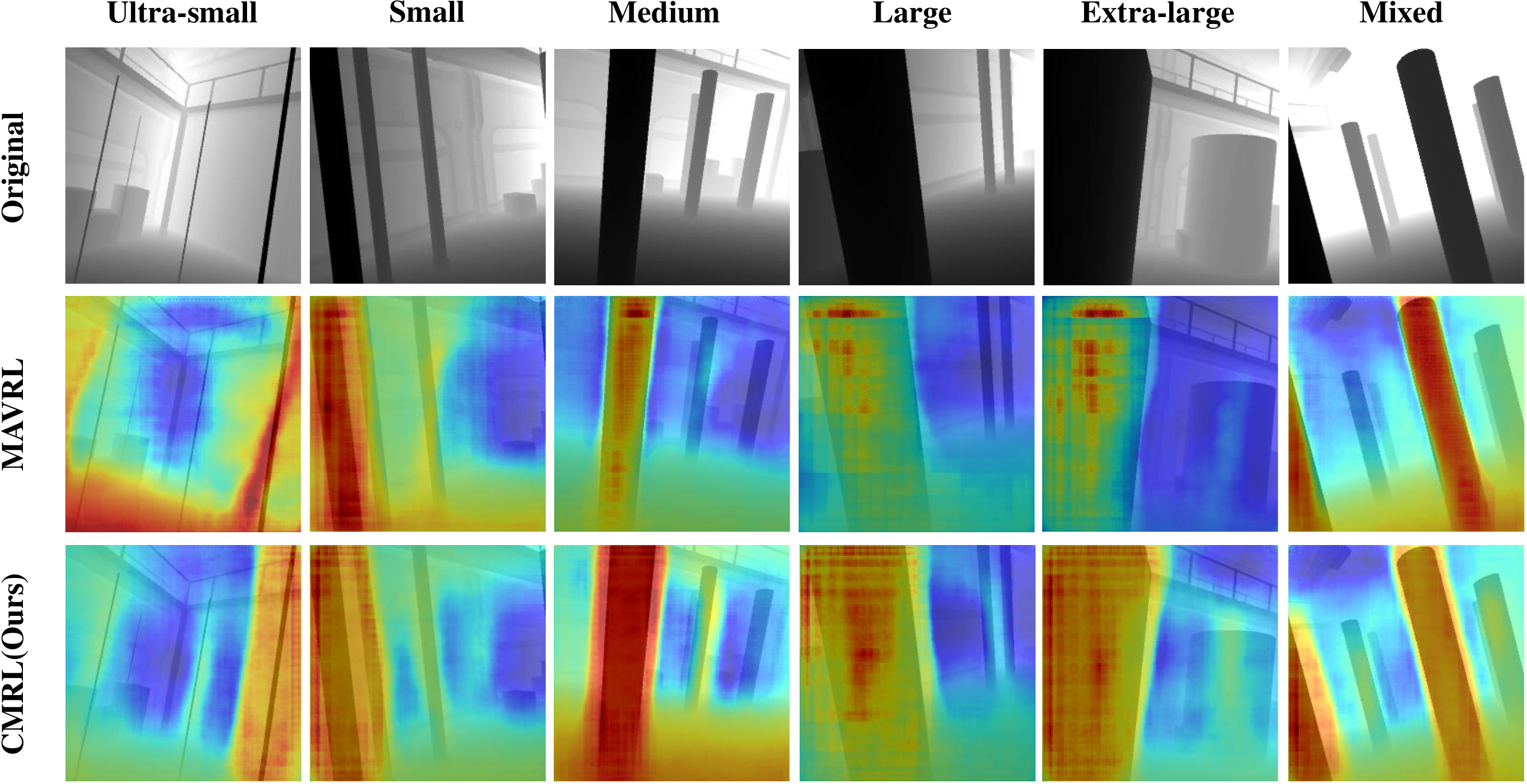}
  \caption{Grad-CAM attention visualizations of MAVRL and CMRL across obstacle scales.}
  \label{fig:attention_vis}
\end{figure*}

Averaging over the radius range can hide how each method responds to increasing clutter.
For the three most demanding environments, we therefore report the result at each individual radius.
Fig.~\ref{fig:box_density} shows the success rate distributions over the eight repetitions.

In the ultra-small environment of Fig.~\ref{fig:box_density}(a), neither baseline is consistently better than the other, and the two stay within 0.07 of each other at every density.
This pattern indicates that their failures come from missing the thin structures themselves rather than from clutter.
CMRL stays at or above 0.72 in all four settings.

The two baselines separate clearly in the extra-large environment of Fig.~\ref{fig:box_density}(b).
MAVRL benefits the most from sparser layouts, rising from 0.33 at 6.0 m to 0.66 at 7.0 m, while CMRL is almost flat from 0.74 to 0.83.
Large obstacles occlude most of the field of view once they are placed close together, so the policy has to act on a heavily truncated observation.
The small degradation of CMRL suggests that the memory module retains enough spatial context to keep detouring decisions consistent.

The mixed-scale environment of Fig.~\ref{fig:box_density}(c) is the most sensitive to density for all three methods.
MAVRL falls from 0.81 at the sparsest setting to 0.43 at the densest, while CMRL only falls from 0.95 to 0.72.
The gap between the two methods therefore more than doubles, so a comparison at a single density can understate the difference between methods.

Across the three environments, CMRL shows the smallest degradation from the sparsest to the densest layout, with 0.12, 0.09, and 0.23 against 0.19, 0.33, and 0.38 for MAVRL.
The standard deviation of CMRL stays between 0.04 and 0.07 in all settings, so its higher success rate does not come with less consistent behavior.
These results indicate that the advantage of CMRL is not tied to a particular clutter level, and that it degrades more gracefully as the environment becomes harder.

\subsection{Network Attention Visualization}

To qualitatively examine the visual regions emphasized by the learned policy, we use Grad-CAM \citep{selvaraju2017grad} to visualize the contribution of different spatial locations in the input depth image.

We apply Grad-CAM to scenes with obstacles at different scales and compare CMRL with MAVRL, as shown in Fig.~\ref{fig:attention_vis}.
In scenes containing small-scale and medium-scale obstacles, CMRL tends to assign stronger attention to small obstacles and potential risk regions at longer distances,
whereas MAVRL often exhibits weaker or more diffuse responses to these structures.
In scenes with large-scale obstacles, CMRL shows more concentrated responses on obstacle bodies and contours, while MAVRL distributes attention more broadly over both obstacle and background regions.
CMRL also tends to highlight surrounding boundary regions near obstacles, suggesting improved sensitivity to safety margins.
These qualitative patterns are consistent with the superior navigation performance of CMRL in multi-scale environments.
\begin{figure}[pos=t]
  \centering
  \includegraphics[width=0.95\columnwidth]{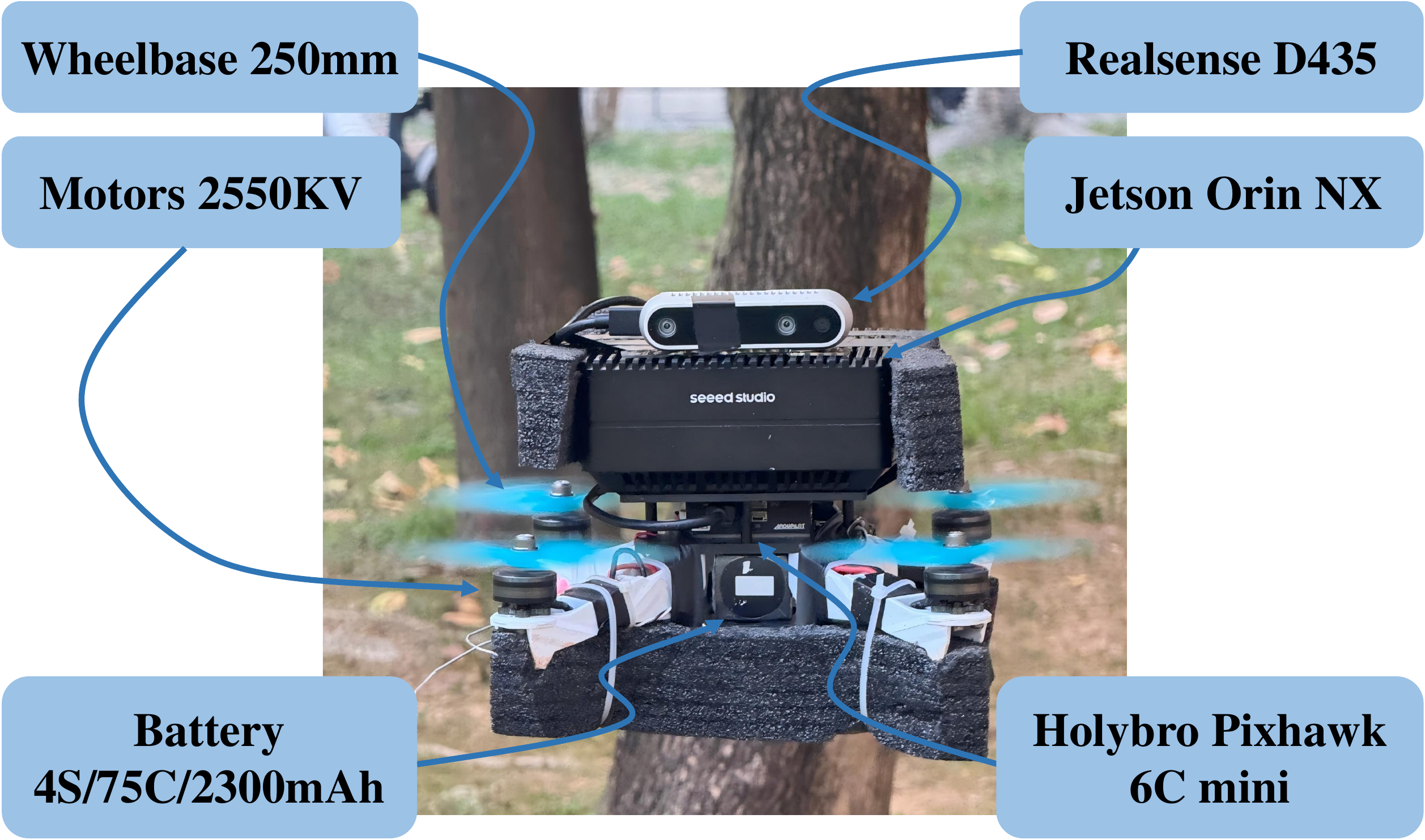}
  \caption{The quadrotor platform used for the real-world experiments.}
  \label{fig:platform}
\end{figure}

\begin{figure*}[pos=b]
  \centering
  \includegraphics[width=0.95\textwidth]{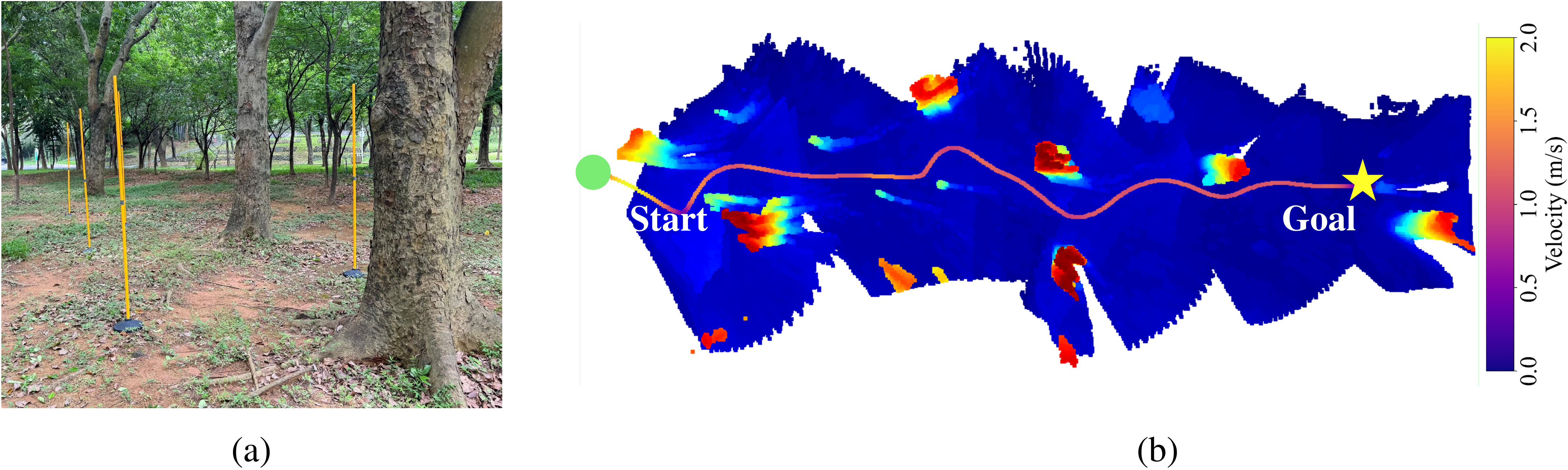}
  \caption{Real-world outdoor experiments.
  (a) The dense forest test environment with multi-scale obstacles.
  (b) A representative flight trajectory from the start to the goal over the reconstructed scene.}
  \label{fig:outdoor_flight}
\end{figure*}

\subsection{Real-World Tests}
To examine the deployability of CMRL in outdoor scenarios, we deploy the method on a quadrotor with a 250 mm wheelbase, as shown in Fig.~\ref{fig:platform}.
The platform is equipped with an Intel RealSense D435 depth camera for depth sensing, an NVIDIA Jetson Orin NX for onboard computation, and a Pixhawk 6C Mini flight controller for low-level actuation.
Although the Jetson Orin NX can achieve an inference frequency of 15 Hz, we set the policy to run at 10 Hz in both simulation and reality.
This preserves onboard computing power for data logging and the MPC module.
Simultaneously, the MPC runs at 100 Hz to generate low-level body-rate and thrust commands.
Outdoor position estimation is provided by a visual-inertial odometry system.

To bridge the sim-to-real gap, we align the depth camera resolution and field of view with the simulation setup.
We collect approximately 5,000 real-world depth images from the outdoor forest scenes and fine-tune the VAE and LSTM representation modules at a reduced learning rate, following the same training procedure as in Section~\ref{sec:representation}.
The policy network trained in simulation is kept frozen throughout, and its outputs are converted to executable commands via the Agilicious MPC controller.

We conduct flight tests in a dense forest environment, shown in Fig.~\ref{fig:outdoor_flight}(a).
The quadrotor is tasked with reaching a designated goal while avoiding obstacles.
To further examine the multi-scale obstacle avoidance capability of CMRL in the real world, we additionally place several slender poles with a diameter of about 2 cm among the surrounding tree trunks.
The thick tree trunks and the slender poles together form an environment that contains obstacles at clearly different scales.
Fig.~\ref{fig:outdoor_flight}(b) shows a representative flight trajectory.
The quadrotor successfully traverses both the slender poles and the thick tree trunks and finally reaches the goal.
The average flight speed is approximately 1.4 m/s.
A video of the outdoor flight tests is available at \url{https://honghongdev.github.io/camerl/}.
These results qualitatively demonstrate that CMRL can be deployed on a real quadrotor platform for multi-scale obstacle avoidance in challenging outdoor scenarios with onboard sensing and computation.

\section{Conclusion}
\label{sec:conclusion}
This paper presents CMRL, a collision-aware and memory-enhanced reinforcement learning framework for UAV navigation in multi-scale obstacle environments.
CMRL combines collision-aware representation learning from depth observations with temporal memory modeling.
This design improves obstacle avoidance performance across obstacle scales, especially in challenging ultra-small and extra-large settings.
Simulation results show that CMRL consistently outperforms baseline methods in success rate while maintaining competitive flight efficiency in multi-scale environments.
Real-world outdoor flight tests further provide qualitative evidence that CMRL can be deployed for obstacle avoidance and goal-reaching with onboard sensing and computation.
The same policy is used at every obstacle scale without retraining, and it runs onboard at 10 Hz on an embedded computer.
CMRL can therefore serve as a perception and decision component for inspection and survey applications in which obstacle sizes differ widely.
Future work will extend the framework to dynamic obstacles and more complex real-world environments.

%%%%%%%%%%%%%%%%%%%%%%%%%%%%%%%%%%%%%%%%%%%%%%%%%%%%%%%%%%%%%%%%%%%%%%%%%%%%%%%%
\ifanon
\else
\section*{CRediT authorship contribution statement}

\textbf{Hong Hong}: Conceptualization, Methodology, Software, Visualization, Writing -- original draft.
\textbf{Feiyu Liao}: Software, Visualization, Writing -- review \& editing.
\textbf{Yongheng Liang}: Supervision.
\textbf{Boning Zhang}: Validation, Data curation.
\textbf{Michael Kunyu Wang}: Validation, Data curation.
\textbf{Haitao Wang}: Visualization, Supervision, Writing -- review \& editing.
\textbf{Hejun Wu}: Resources, Writing -- review \& editing.
\fi

\ifanon
\else
\section*{Acknowledgments}

This work was mainly supported by the National Natural Science Foundation of China (NSFC) under Grant No. 62272497 (Hejun Wu) and the National Key Research and Development Program of China under Grant 2023YFB\allowbreak 4301900.
\fi

\section*{Data availability}

\ifanon
The training and evaluation code is available at \url{https://anonymous.4open.science/r/artifact-B8E5/}.
The real-world depth image data collected in this study are available from the corresponding author on reasonable request.
\else
The training and evaluation code is publicly available at \url{https://honghongdev.github.io/camerl/}.
The real-world depth image data collected in this study are available from the corresponding author on reasonable request.
\fi

\bibliographystyle{cas-model2-names}

\bibliography{cmrl}

\end{document}